\documentclass[twocolumn]{article}
\usepackage{graphicx} 
\usepackage{xcolor}
\usepackage{indentfirst}
\usepackage{float}
\usepackage{algorithm}
\usepackage{algpseudocode}
\usepackage{amsmath}
\usepackage[utf8]{inputenc}
\usepackage{hyperref} 
\usepackage{natbib}

\usepackage{amssymb}
\usepackage{booktabs} 
\usepackage{array}    
\title{A Different Level Text Protection Mechanism With Differential Privacy}
\author{Qingwen Fu}

\begin{document}

\maketitle
    
\section{Abstract}

With the widespread application of differential privacy in text protection, however, the current text cleaning mechanism based on metric local differential privacy (MLDP) is not applicable to non-metric semantic similarity measurement, and cannot achieve a good trade-off between privacy and practicality. And currently when we perform differential privacy on long text data, all text data will be perturbed. This method of perturbing all texts may be relatively effective for downstream tasks on some data sets, but if applied to long text data, it may have a great impact on the overall meaning of the text. Therefore, in this article, we propose to use the weights of different words in the pre-trained model to assign different weight parameters to words of different importance. Perform differential perturbations. In addition to conducting inference attacks, we also use large models to perform privacy and validity tests on our perturbed data.
\section{Introduction}

In many natural language processing (NLP) applications, input text often contains sensitive information that can infer the identity of a specific person \cite{jegorova2022survey}. In addition, legal restrictions such as CCPA and GDPR may further restrict the sharing of sensitive text data. This makes it difficult for NLP service providers to collect training data unless the privacy concerns of data owners (including individuals and institutions) are properly addressed.

A lot of work has been done to address privacy issues \cite{lyu2020differentially,anil2021large,dupuy2022efficient,li2021large} to train language models using differential privacy (DP)\cite{dwork2006calibrating} , which is considered the standard for privacy-preserving computing. These methods protect the data source by adding noise to the gradient or training data. However, they require service providers to collect raw data for LM training, which may still cause privacy leakage.

In order to fundamentally solve the privacy leakage problem, data needs to be fundamentally protected. Typically, these privacy mechanisms \cite{feyisetan2019leveraging, feyisetan2020privacy, yue2021differential} work by replacing the original tokens in the original document with new tokens extracted from the output token set. To generate a cleaned text document. Specifically, they adopt metric local differential privacy (MLDP, also known as \( d\chi \)-privacy) to provide privacy and practicality guarantees. MLDP\cite{chatzikokolakis2013broadening} inherits the idea of DP and ensures that the output of any adjacent input tokens is indistinguishable to protect the original tokens from being inferred. On the other hand, MLDP preserves the utility of the purified text by assigning higher sampling probabilities to tokens that are semantically closer to the original tokens. In these mechanisms, any metric distance (such as Euclidean distance) can be used to measure the semantic similarity between tokens.

In the paper  \cite{chen2022customized}", an MLDP-based concept is proposed to assign a smaller custom output set to each input token to achieve token-level privacy protection. This method is an improvement on the santext\cite{yue2021differential} method, which increases the text perturbation rate without reducing the privacy protection effect by limiting the size of the output set. The custom parameter K can be adjusted to determine the output set size of each input token to achieve different utility-privacy trade-offs, and an improved CusText+ mechanism is proposed to skip stop words when sampling to achieve higher utility.
This analysis does improve the perturbation efficiency of words in the text to a certain extent,
but according to all previous studies, they treat every token that appears in the text equally, which actually perturbs all tokens in the text equally, which may not cause much performance impact on datasets for specific tasks. However, it will cause loss of meaning in common long texts, especially in some medical datasets or long text novels. If all words are treated equally and deemed equally important, and are perturbed to the same extent, this will greatly affect the effectiveness of the text data and lose some of the information we need. Therefore, we propose a method based on a pre-trained BERT model. Using the BERT pre-trained model, the attention weights of all tokens in the sample are extracted, and then the weights of the multi-head multi-layer Transformer are averaged and regularized. This regularized weight is used to symbolically represent the importance of each word in the sample. According to this importance parameter, words of different importance are selectively perturbed. This can reduce the damage to the effectiveness of the text to a certain extent. We tested it on two public datasets, SST-2 and QNLI, and proved the effectiveness of our method of extracting words of different importance.

\section{Related Work}

When discussing privacy risks and protection measures in natural language processing (NLP), we can see three main research directions: research on privacy attacks on deep learning models, differential privacy (DP) and its application in NLP, and the application of local differential privacy (LDP).

First, privacy attacks against deep learning models, especially language models (LMs), have become an important research area. For example, \cite{song2020information} proposed a classification for recovering sensitive attributes or parts of original text from text embeddings output by popular LMs without relying on the structure or pattern of the input text. \cite{carlini2021extracting} demonstrated a black-box attack against GPT-2, capable of extracting verbatim text of the training data. These studies show that privacy attacks on LMs are realistic and damaging, so it is crucial to develop defenses with strict safeguards.

Secondly, in terms of Differential Privacy (DP) and its application in NLP, DP has become the de facto standard for statistical analysis. For example, some research attempts to inject high-dimensional DP noise into text representations \cite{feyisetan2019leveraging,feyisetan2020privacy,xu2020differentially} but these methods fail to achieve a good balance between privacy and utility, mainly because of the “dimensionality Curse". Another approach is to learn private text representations through adversarial training \cite{xie2017controllable,coavoux2018privacy}, where the adversary model is trained to infer sensitive information together with the master model , while the master model is trained to maximize the adversary's loss and minimize the main learning objective.

Third, the application of local differential privacy (LDP) also plays an important role in NLP. LDP allows data owners to sanitize data locally before sending it to the server. This means data owners can share information without revealing the content of their original data. In NLP applications, LDP is particularly valuable because it can collect and analyze text data while protecting user privacy. For example, the LDP mechanism can be used to generate sanitized text datasets that can be used to train machine learning models without exposing personal information. The challenge of LDP is to achieve privacy protection while maintaining data practicality, especially when dealing with text data with complex structure and high-dimensional features.

To sum up, the NLP field faces multiple challenges when dealing with privacy protection issues. On the one hand, effective defense strategies need to be developed against privacy attacks on LMs; on the other hand, differential privacy and local differential privacy provide a series of solutions to protect the privacy of text data. These studies not only help improve the privacy protection capabilities of existing technologies, but also provide important guidance for future privacy protection research in the field of NLP.

\section{Preliminaries}

Before we delve deeper into our CusText technique, let's first briefly review some fundamental concepts, including \(\epsilon\)-differential privacy and the exponential mechanism.

{Definition 1 (\(\epsilon\)-differential privacy)}
Given a privacy parameter \(\epsilon \geq 0\), for all adjacent input pairs \(x, x' \in X\), and for every possible output \(y \in Y\), a randomized mechanism \(M\) satisfies \(\epsilon\)-differential privacy if it adheres to the following condition:
\[
\frac{\Pr[M(x) = y]}{\Pr[M(x' ) = y]} \leq e^\epsilon
\]
In this definition, a smaller \(\epsilon\) indicates a higher level of privacy protection. Theoretically, \(\epsilon\)-DP ensures that even adversaries with infinite computing power cannot distinguish between the probability distributions of two adjacent inputs, as their probabilities of producing the same output \(y\) are closely matched. In the context of Natural Language Processing (NLP), any pair of input tokens that produce the same output set \(Y\) are considered adjacent. This paper continues to use this definition for adjacent inputs.

\textbf{Definition 2 (Exponential Mechanism)}.
Given a scoring function $u: X \times Y \rightarrow \mathbb{R}$, the exponential mechanism $M(X, u, Y)$ achieves $\epsilon$-differential privacy by randomly selecting an output token $y \in Y$ to perturb the input token $x \in X$ with a probability proportional to 
\[
e^{\frac{\epsilon \cdot u(x,y)}{2\Delta u}}
\]
Here, $u(x, y)$ represents the score of the output token $y$ for the input token $x$. Additionally, the sensitivity of $u$, denoted as $\Delta u$, for the exponential mechanism (EM) is defined by 
\[
\Delta u := \max_{y \in Y} \max_{x, x' \in X} \left| u(x, y) - u(x', y) \right|
\]

According to the second definition, lower sensitivity makes it statistically more difficult to distinguish the original token from its adjacent tokens. In practice, we may standardize the scoring function \(u\), normalizing its sensitivity \(\Delta u\) to a fixed value (e.g., 1), so that the selection probability for each output token \(y\) for an input token \(x\) is solely related to \(u(x, y)\), considering that \(\epsilon\) and \(\Delta u\) are predetermined, and a larger \(u(x, y)\) results in a higher sampling probability.

In an NLP task, we assume each document \(D = \langle R_i \rangle_{i=1}^m\) contains \(m\) records, and each record \(R = \langle t_j \rangle_{j=1}^n\) contains \(n\) tokens. We define the task of text sanitization as follows: Given an input document \(D\) containing sensitive information, a set of all possible input tokens \(X\), a set of all possible output tokens \(Y\), and a differential privacy mechanism \(M\) (e.g., the EM used in this work), it applies the mechanism \(M\) to each input token \(t_j \in D\), replacing it with an output token \(t'_j \in Y\) if \(t_j \in X\). All tokens after replacement form the sanitized document, i.e., \(D' = \langle R'_i \rangle_{i=1}^m\) and \(R' = \langle t'_j \rangle_{j=1}^n\).

Following previous studies \cite{xu2020differentially,feyisetan2019leveraging,yue2021differential,chen2022customized,qu2021natural}, we still adopt a semi-honest threat model in the context of local differential privacy. In this model, the data owner only submits sanitized documents to the service provider. However, a malicious service provider may try to extract sensitive information from the received data. We assume that the adversary can only obtain the sanitized text and all algorithms and mechanisms are public and transparent. In addition, we also assume that the adversary has unlimited computing power.

\section{Method}
Our privacy perturbation method is based on the CusText mechanism. The difference is that we use the BERT pre-trained model to assign weights to different words in the same example. Then we average the weights of multiple heads and layers. We remove the weights of CLS and septoken and regularize the weights of other words. Use this weight value to represent the importance of different words. Then we combine the CusText mechanism to perform different degrees of perturbation for our words of different importance.

"CusText" is a tailored text sanitization framework designed to safeguard privacy by substituting every token within a text. It comprises two primary components: firstly, a semantic correlation-based mapping function, fmap, which identifies the appropriate output set for each input token; secondly, a sampling function, fsample, that selects new tokens from this output set using an exponential mechanism.

Unlike traditional SANTEXT methods, CusText enhances the relevance of the output tokens to the original tokens by customizing the output set for each input token, thus improving the utility of the model. The development of the mapping function involves picking tokens from the input set, identifying those that are semantically closest, and creating a mapping. This mapping is then refined by progressively removing the tokens that have been mapped until a complete mapping is achieved or there are insufficient tokens left to continue. This strategy ensures that every input token is paired with at least one neighboring token, preserving the effectiveness of the privacy measures.

\begin{algorithm}
\caption{CusText Mapping Mechanism}
\begin{algorithmic}[1] 
\State \textbf{Input:} Customization parameter \(K\), input set \(X\), output set \(Y = X\), similarity measure \(d\)
\State \textbf{Output:} Mapping Function \(f_{\text{map}}\)
\While{\(|X| \geq K\)} 
    \State Pick an arbitrary token \(x\) from \(X\)
    \State Initialize an output set \(Y' = \{x\}\) for \(x\)
    \ForAll{$y \in Y \setminus \{x\}$} 
        \State Compute the similarity \(d(x, y)\) of \(x\) and \(y\)
    \EndFor
    \State Add the top-\((K-1)\) tokens that are semantically closest to \(x\) to \(Y'\) based on \(d(., .)\)
    \ForAll{$x' \in Y'$} 
        \State Assign the output set of \(x'\) as \(Y'\)
    \EndFor
    \State Update \(X \leftarrow X \setminus Y'\) and \(Y \leftarrow Y \setminus Y'\) 
\EndWhile
\State Perform Lines 2--9 for the remaining tokens in \(X\) and \(Y\) with customization parameter \(K' = |X|\)
\State \textbf{return} \(f_{\text{map}}\)
\end{algorithmic}
\end{algorithm}

Sampling function: The fsample function, which is reliant on the fmap, selects an output tag for each input tag. This selection is governed by an exponential mechanism, and it requires a carefully designed scoring function u to maintain a balance between utility and privacy. The function ensures that the relationship between each input and output tag pair is capped, with pairs that are semantically closer receiving higher scores.

Scoring Function,custext is based on the same similarity function used in mapping schemes,,e.g., Euclidean distance or cosine similarity based on token-vector,representations \cite{mikolov2013efficient,pennington2014glove}.,In general, all similarity measures can be divided into two categories,,negative and positive,,according to the correlation between the score and semantic proximity.,For example, Euclidean distance and cosine similarity are negative,and positive correlation measures, respectively, because the smaller the Euclidean distance,and the larger the cosine value between two vectors,,means that the semantic proximity of their corresponding tokens is higher.,Next, we will design scoring functions for these two types of similarity,measurements.

The following is our own perturbation method based on words of different importance. Our method mainly uses the pre-extracted words of different importance as our sensitive word list, and then uses the custext method to perturb these sensitive word lists.

\begin{algorithm}
\caption{Different levels of protection mechanism}
\begin{algorithmic}[1] 
\State \textbf{Input:} Original document \(D = (R_i)_{i=1}^m\), sampling function \(f_{\text{sample}}\), different level sensitive word list \(S\)
\State \textbf{Output:} different level sanitized document \(D'\)
\State Initialize the sanitized document \(D' = \emptyset\)
\For{all record \(R \in D\)}
    \State Initialize the sanitized record \(R' = \emptyset\)
    \For{all token \(x \in R\)}
        \If{token is used and \(x \in S\)}
            \State \(x' \leftarrow f_{\text{sample}}(x)\) and append to \(R'\)
        \Else
           \State Append \(x\) to \(R'\)
        \EndIf
    \EndFor
    \State Add \(R'\) to \(D'\)
\EndFor
\State \textbf{return} \(D'\)
\end{algorithmic}
\end{algorithm}
Aggressive mechanism.
When we select the important vocabulary list, if we adopt an aggressive mechanism, we can perturb all the words in the sensitive vocabulary list without difference, but this may have a greater impact on the original semantics of the text, because the same noun or the same verb will be perturbed into different words, which will cause the text semantics to be incoherent. The result for short text may be less than the effect on long text.

Conservative mechanism.
When the same sensitive word appears multiple times in a sample, we give it the same perturbation result. This is a conservative mechanism and may be easier to attack. But it is possible to give the same nouns the same perturbation in the content of long texts. In this way, the relationship between words such as subject and predicate can be better preserved, and its semantic structure can be preserved. It is possible to protect sensitive information while still having better semantic information and text information.
The above two mechanisms can be used to process different categories of text data, and can be freely selected as needed. Combined with our selection mechanism for words of different degrees of importance, the text can be protected more flexibly to better achieve a balance between privacy and utility.

\section{Experiment}.
Experimental Setup,

5.1 Experimental Setup
Following \cite{feyisetan2020privacy,yue2021differential}
We selected two datasets from the GLUE benchmark \cite{wang2018glue} in our experiments, both of which contain
In our experimental section, we aim to demonstrate the efficacy of using attention mechanism parameters to represent the importance of different words within a sample. This section is divided into two parts, each utilizing the public datasets SST-2 and QNLI to validate our method.

\textbf{Datasets Description:}
\begin{itemize}
    \item \textbf{SST-2:} A widely-used movie review dataset for sentiment classification, consisting of 67,000 training samples and 1,800 test samples. The evaluation metric is accuracy.
    \item \textbf{QNLI:} A dataset for sentence pair classification with 105,000 training samples and 5,200 test samples. Accuracy is also used as the evaluation metric here.
\end{itemize}

In our approach, for both the SST-2 and QNLI datasets, we first identify the most and least important words, quantified as the top and bottom 10\%, 20\%, 30\%, 40\%, 50\%, and 60\% based on the attention scores. These words are considered as the sensitive words that need to be perturbed. We record the number of words actually perturbed during training and compare it under similar total perturbation conditions to gauge the effectiveness of our method.
We use the vocabulary from CounterFitting in GloVe, and apply both Euclidean distance and cosine similarity as measures for comparing GloVe vectors. The sensitive word list is derived from the probabilities associated with different words in the pre-trained model.
For each downstream task, we set the maximum sequence length to $128$ and limit the training to $3$ epochs. On both SST-2 and QNLI datasets, the batch size is set to $64$. We use \texttt{bert-base-uncased} as the pre-trained model with an increased learning rate of $2 \times 10^{-5}$. The experiments are conducted on an A100 GPU.

The second part of our experimental analysis focuses on demonstrating the effectiveness of our approach. In this phase, we perturb words of varying degrees of importance—specifically, 5\%, 10\%, and 20\% of the words determined by our quantifier. We then evaluate both the privacy and effectiveness of the perturbed datasets using several established mechanisms.

\begin{itemize}
    \item \textbf{Evaluation Mechanisms:} We apply various metrics to assess the privacy levels and the utility of the datasets after perturbation.
    \item \textbf{Data Perturbation:} We methodically perturb the words identified as having high, medium, and low importance to measure the impact on the dataset’s utility and privacy.
    \item \textbf{Analysis of Important Words:} This method also allows us to count and calculate the distribution of words based on their importance. We identify and examine some relatively high-importance words, observe the categories they belong to, and analyze their patterns.
\end{itemize}

This structured evaluation helps in understanding how different levels of perturbation affect the privacy-security balance and the overall effectiveness of the sensitive data we intend to protect.

\subsection{Experiment Result}
Below are some of my experimental results when \(\epsilon\) equals to 3.

\begin{figure}[H]
    \centering
    \begin{minipage}{0.45\textwidth}
        \includegraphics[width=\linewidth]{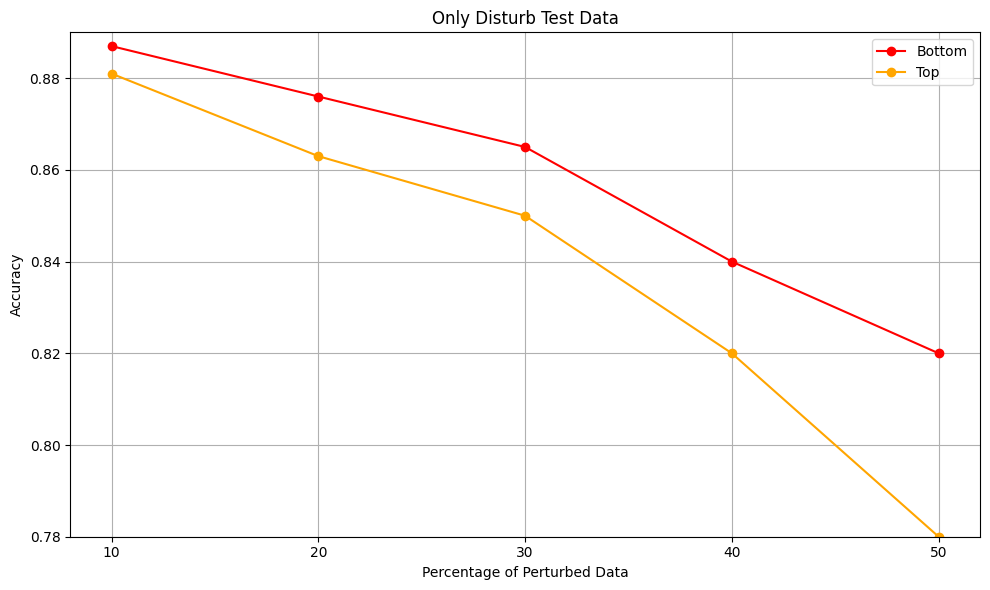 }
        \caption{Only Disturb Test Data for SST2}
    \end{minipage}
    \hfill
    \begin{minipage}{0.45\textwidth}
        \includegraphics[width=\linewidth]{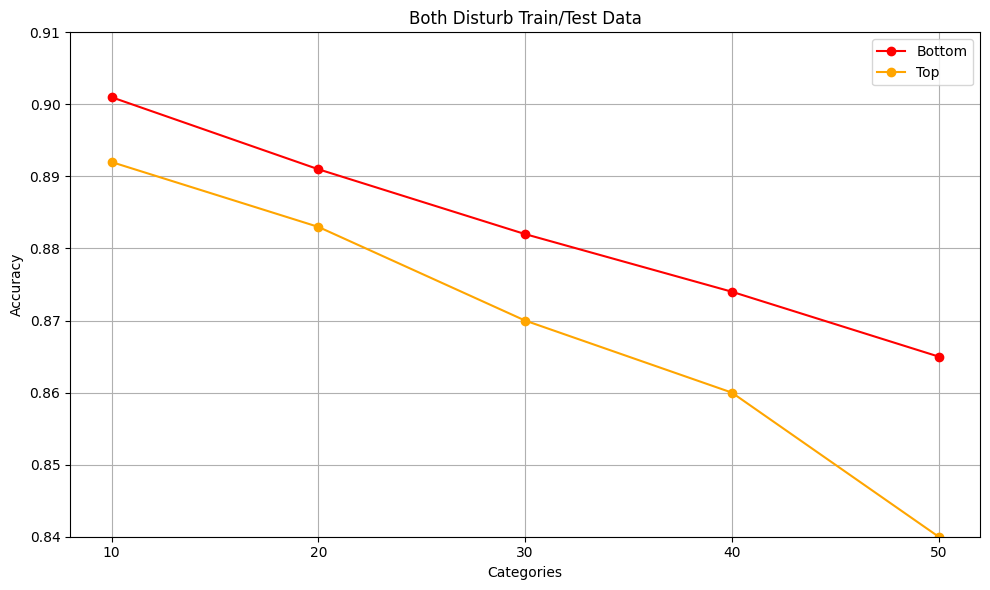 }
        \caption{Both Disturb Train and Test Data for SST2}
    \end{minipage}
\end{figure}

\begin{figure}[H]
    \centering
    \begin{minipage}{0.45\textwidth}
        \includegraphics[width=\linewidth]{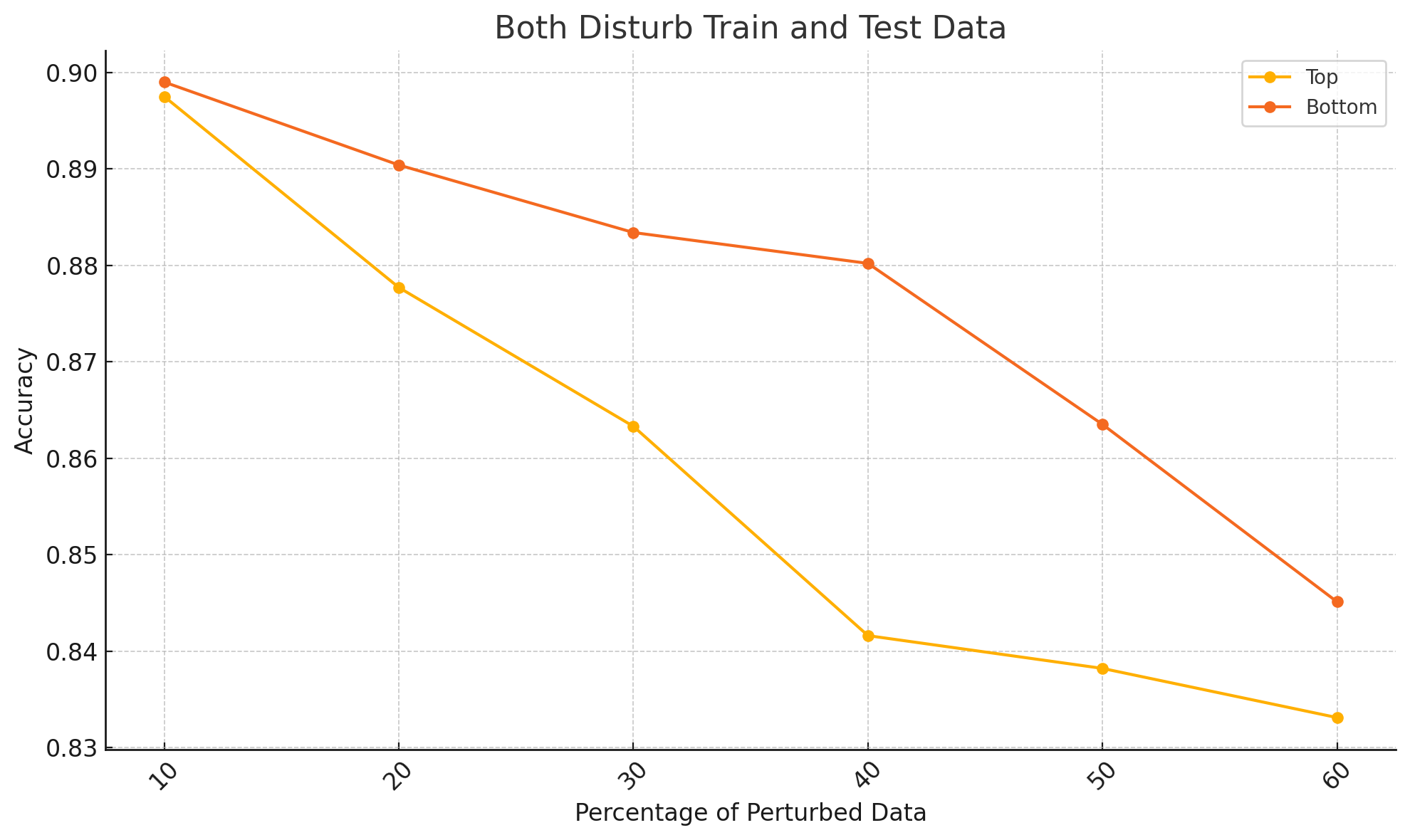}
        \caption{Only Disturb Test Data for QNLI}
        \label{fig:only_test_data}
    \end{minipage}
    \hfill
    \begin{minipage}{0.45\textwidth}
        \includegraphics[width=\linewidth]{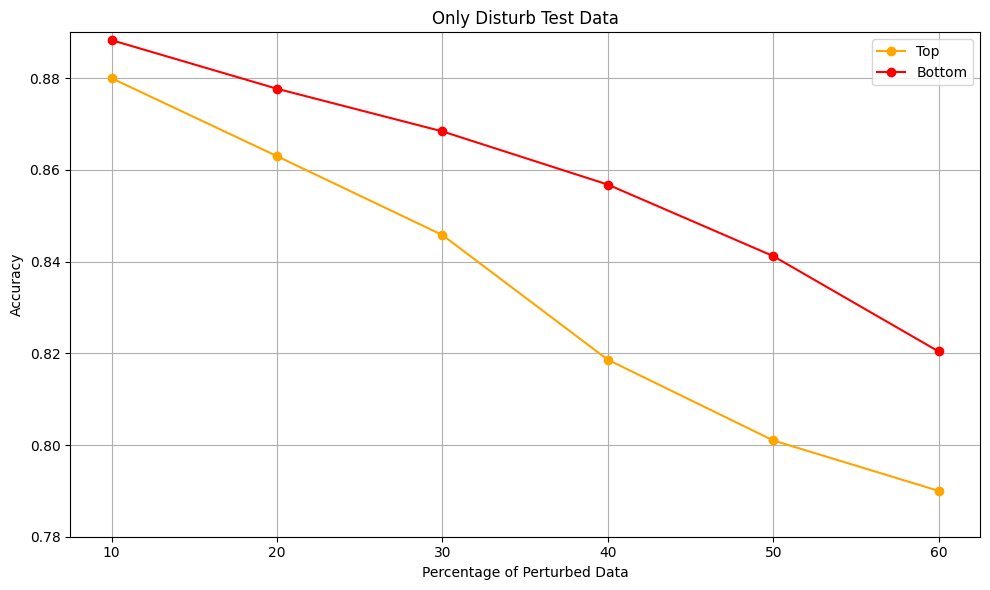}
        \caption{Both Disturb Train and Test Data for QNLIs}
        \label{fig:both_disturb}
    \end{minipage}  
\end{figure}

\begin{table}[H]
\renewcommand{\arraystretch}{1.2} 
\centering
\begin{tabular}{>{\raggedright\arraybackslash}m{3cm} >{\centering\arraybackslash}m{2cm} >{\centering\arraybackslash}m{2cm}}
\toprule
\textbf{SST3 Dataset} & \textbf{Train/Test Accuracy} & \textbf{Only Test Data Accuracy} \\
\midrule
Top 50    & 0.840 & 0.780 \\
Top 40    & 0.860 & 0.820 \\
Top 30    & 0.870 & 0.850 \\
Top 20    & 0.883 & 0.863 \\
Top 10    & 0.892 & 0.881 \\
Bottom 50 & 0.865 & 0.820 \\
Bottom 40 & 0.874 & 0.840 \\
Bottom 30 & 0.882 & 0.865 \\
Bottom 20 & 0.891 & 0.876 \\
Bottom 10 & 0.901 & 0.887 \\
No Disturbance & 0.905 & -    \\
Full Disturbance & 0.72 & -  \\
\bottomrule
\end{tabular}
\caption{Accuracy results for the QNLI dataset under different conditions.}
\end{table}
\begin{table}[H]
\centering
\begin{tabular}{>{\raggedright\arraybackslash}m{3cm} >{\centering\arraybackslash}m{2cm} >{\centering\arraybackslash}m{2cm}}
\toprule
\textbf{QNLI Dataset} & \textbf{Train/Test Accuracy} & \textbf{Only Test Data Accuracy} \\
\midrule
Top 60      & 0.8331 & 0.79   \\
Bottom 60   & 0.8451 & 0.8204 \\
Top 50      & 0.8382 & 0.801  \\
Bottom 50   & 0.8635 & 0.8412 \\
Top 40      & 0.8416 & 0.8186 \\
Bottom 40   & 0.8802 & 0.8568 \\
Top 30      & 0.8633 & 0.8458 \\
Bottom 30   & 0.8834 & 0.8684 \\
Top 20      & 0.8777 & 0.8656 \\
Bottom 20   & 0.8904 & 0.8777 \\
Top 10      & 0.8975 & 0.88   \\
Bottom 10   & 0.899  & 0.8864 \\
No Disturbance & 0.9096 & -    \\
Full Disturbance & 0.7133 & -  \\
\bottomrule
\end{tabular}
\caption{Accuracy results for the QNLI dataset under different conditions.}
\label{tab:qnli-results}
\end{table}
\begin{table}[H]
    \centering
    \begin{tabular}{>{\raggedright\arraybackslash}p{3cm} >{\centering\arraybackslash}p{2cm} >{\centering\arraybackslash}p{2cm}}
        \toprule
        Top N Accuracy & Conservative Strategy & Aggressive Strategy \\
        \midrule
        Top 10 & 0.901 & 0.88 \\
        Top 20 & 0.8864 & 0.8656 \\
        Top 30 & 0.8756 & 0.8458 \\
        Top 40 & 0.8623 & 0.8186 \\
        Top 50 & 0.8428 & 0.801 \\
        Top 60 & 0.821 & 0.79 \\
        \bottomrule
    \end{tabular}
    \caption{Accuracy comparison between conservative and aggressive strategies}
    \label{tab:accuracy_comparison}
\end{table}

\subsection{result analysis}\
For the perturbation of data with different importance, we conducted experiments on the SST-2 and QNLI datasets. For fair comparison in the future, we chose Glove as the token embedding and controlled other variables to be the same. Table 1 shows the results of perturbing words of different importance on the SST-2 dataset while keeping the training set unchanged. For the same test set, words of different importance are perturbed while keeping $\epsilon = 3$ unchanged. As can be seen from the figure, when the test set data is perturbed with basically the same amount of data, the result of perturbing more important words is worse than that of perturbing less important words, which also proves that our vocabulary extraction method is correct. Figure 2 shows the results of perturbing the training data and test data at the same time. The results show that when perturbing the same number of words of different importance, perturbing more important words has a greater impact on the results, which also proves that our extraction strategy is correct. When we make a horizontal comparison, we find that when we use the perturbed training set for training, the matching effect with the test set is better, which also reflects the effectiveness of our method for words of different importance to a certain extent. When we observe the results of the QNLI dataset, we can also draw the above conclusions. Therefore, our Transformer-based extraction method is effective. When we perform differential privacy on the text, we can selectively perturb words of different importance. Of course, this method can also be used as a screening mechanism to help us narrow the search scope of keywords and privacy words. Combined with other named entity recognition and LLM reasoning methods, it can help us find more effective keywords faster. This will be a general method.

Conservative method, when we adopt a conservative strategy, that is, when we keep the same perturbation results for the same words in the same sample, the results are as follows. Here we mainly analyze the results of the guard strategy for the top 10, 20, 30, 40, and 50 emphasized words of qnli (long text length, with a greater possibility of the same vocabulary). It can be observed that when we consider adopting the guard strategy, the results of the experiment will be significantly improved. Therefore, we can use the guard strategy in the relatively long policy text protection process, which can better maintain semantic coherence.

Token reasoning attacks and query attacks are carried out on the perturbed text to test the effectiveness of our extraction of data of different importance and its relevance to privacy. Using a pre-trained BERT model can help infer the possibility of recovering the original text from the purified text. By replacing each token in the purified text with the "[MASK]" token and inputting it into the BERT model, we can get the model's predicted output for "[MASK]", which is the inferred original token. If the predicted output is the same as the token of the original input, we consider the attack attempt to be successful. By calculating the success rate of all such attacks (rmask), we can measure the privacy protection of the text, which is 1-rmask. Because our algorithm is based on the custext algorithm and has not been modified to the original algorithm, its effect is the same as custext.

Importance vocabulary analysis When we use chatgpt4 to analyze the words of different importance we extracted, we find that the more important words are often those nouns, pronouns, punctuation marks, etc. This is the same as the more important words in a sentence we understand. However, when we use GPT4 and our more important words to reconstruct the zero-shot sentence, the reconstructed sentence is very different from our original sentence. Therefore, our method does not perform well under unguided reconstruction. This method may be more suitable for identifying important words.

\section{Conclusion and limitation}
Conclusion: This method proves that we can reflect the importance of different words in different sentences through multiple layers of Transformers and the attention weights between them, but more supplementary experiments are needed. Moreover, when we apply this method to long text data, our accuracy will be biased due to the limitation of the maximum length of Transformer and the long text length. This requires us to combine some other models and find a way to obtain longer length data at the same time. We need to do more work on this basis to improve its performance. We can do more experiments and research on this basis in combination with LLM.

\section{Future work}
With the recent emergence and development of LLM, I think we can combine the large oracle model with the discovery of sensitive data. Combined with prompts, LLM can identify important and sensitive information in the text. And we can combine LLM with this method to filter sensitive information in the text except for specific categories, because some other information in the text that is not classified may also contain some critical sensitive information. This is a direction worth exploring.

\bibliographystyle{apalike} 
\bibliography{reference}

\begin{thebibliography}{}

\bibitem[Anil et~al., 2021]{anil2021large}
Anil, R., Ghazi, B., Gupta, V., Kumar, R., and Manurangsi, P. (2021).
\newblock Large-scale differentially private bert.
\newblock {\em arXiv preprint arXiv:2108.01624}.

\bibitem[Carlini et~al., 2021]{carlini2021extracting}
Carlini, N., Tramer, F., Wallace, E., Jagielski, M., Herbert-Voss, A., Lee, K., Roberts, A., Brown, T., Song, D., Erlingsson, U., et~al. (2021).
\newblock Extracting training data from large language models.
\newblock In {\em 30th USENIX Security Symposium (USENIX Security 21)}, pages 2633--2650.

\bibitem[Chatzikokolakis et~al., 2013]{chatzikokolakis2013broadening}
Chatzikokolakis, K., Andr{\'e}s, M.~E., Bordenabe, N.~E., and Palamidessi, C. (2013).
\newblock Broadening the scope of differential privacy using metrics.
\newblock In {\em Privacy Enhancing Technologies: 13th International Symposium, PETS 2013, Bloomington, IN, USA, July 10-12, 2013. Proceedings 13}, pages 82--102. Springer.

\bibitem[Chen et~al., 2022]{chen2022customized}
Chen, H., Mo, F., Wang, Y., Chen, C., Nie, J.-Y., Wang, C., and Cui, J. (2022).
\newblock A customized text sanitization mechanism with differential privacy.
\newblock {\em arXiv preprint arXiv:2207.01193}.

\bibitem[Coavoux et~al., 2018]{coavoux2018privacy}
Coavoux, M., Narayan, S., and Cohen, S.~B. (2018).
\newblock Privacy-preserving neural representations of text.
\newblock {\em arXiv preprint arXiv:1808.09408}.

\bibitem[Dupuy et~al., 2022]{dupuy2022efficient}
Dupuy, C., Arava, R., Gupta, R., and Rumshisky, A. (2022).
\newblock An efficient dp-sgd mechanism for large scale nlu models.
\newblock In {\em ICASSP 2022-2022 IEEE International Conference on Acoustics, Speech and Signal Processing (ICASSP)}, pages 4118--4122. IEEE.

\bibitem[Dwork et~al., 2006]{dwork2006calibrating}
Dwork, C., McSherry, F., Nissim, K., and Smith, A. (2006).
\newblock Calibrating noise to sensitivity in private data analysis.
\newblock In {\em Theory of Cryptography: Third Theory of Cryptography Conference, TCC 2006, New York, NY, USA, March 4-7, 2006. Proceedings 3}, pages 265--284. Springer.

\bibitem[Feyisetan et~al., 2020]{feyisetan2020privacy}
Feyisetan, O., Balle, B., Drake, T., and Diethe, T. (2020).
\newblock Privacy-and utility-preserving textual analysis via calibrated multivariate perturbations.
\newblock In {\em Proceedings of the 13th international conference on web search and data mining}, pages 178--186.

\bibitem[Feyisetan et~al., 2019]{feyisetan2019leveraging}
Feyisetan, O., Diethe, T., and Drake, T. (2019).
\newblock Leveraging hierarchical representations for preserving privacy and utility in text.
\newblock In {\em 2019 IEEE International Conference on Data Mining (ICDM)}, pages 210--219. IEEE.

\bibitem[Jegorova et~al., 2022]{jegorova2022survey}
Jegorova, M., Kaul, C., Mayor, C., O'Neil, A.~Q., Weir, A., Murray-Smith, R., and Tsaftaris, S.~A. (2022).
\newblock Survey: Leakage and privacy at inference time.
\newblock {\em IEEE Transactions on Pattern Analysis and Machine Intelligence}, 45(7):9090--9108.

\bibitem[Li et~al., 2021]{li2021large}
Li, X., Tramer, F., Liang, P., and Hashimoto, T. (2021).
\newblock Large language models can be strong differentially private learners.
\newblock {\em arXiv preprint arXiv:2110.05679}.

\bibitem[Lyu et~al., 2020]{lyu2020differentially}
Lyu, L., He, X., and Li, Y. (2020).
\newblock Differentially private representation for nlp: Formal guarantee and an empirical study on privacy and fairness.
\newblock {\em arXiv preprint arXiv:2010.01285}.

\bibitem[Mikolov, 2013]{mikolov2013efficient}
Mikolov, T. (2013).
\newblock Efficient estimation of word representations in vector space.
\newblock {\em arXiv preprint arXiv:1301.3781}.

\bibitem[Pennington et~al., 2014]{pennington2014glove}
Pennington, J., Socher, R., and Manning, C.~D. (2014).
\newblock Glove: Global vectors for word representation.
\newblock In {\em Proceedings of the 2014 conference on empirical methods in natural language processing (EMNLP)}, pages 1532--1543.

\bibitem[Qu et~al., 2021]{qu2021natural}
Qu, C., Kong, W., Yang, L., Zhang, M., Bendersky, M., and Najork, M. (2021).
\newblock Natural language understanding with privacy-preserving bert.
\newblock In {\em Proceedings of the 30th ACM International Conference on Information \& Knowledge Management}, pages 1488--1497.

\bibitem[Song and Raghunathan, 2020]{song2020information}
Song, C. and Raghunathan, A. (2020).
\newblock Information leakage in embedding models.
\newblock In {\em Proceedings of the 2020 ACM SIGSAC conference on computer and communications security}, pages 377--390.

\bibitem[Wang, 2018]{wang2018glue}
Wang, A. (2018).
\newblock Glue: A multi-task benchmark and analysis platform for natural language understanding.
\newblock {\em arXiv preprint arXiv:1804.07461}.

\bibitem[Xie et~al., 2017]{xie2017controllable}
Xie, Q., Dai, Z., Du, Y., Hovy, E., and Neubig, G. (2017).
\newblock Controllable invariance through adversarial feature learning.
\newblock {\em Advances in neural information processing systems}, 30.

\bibitem[Xu et~al., 2020]{xu2020differentially}
Xu, Z., Aggarwal, A., Feyisetan, O., and Teissier, N. (2020).
\newblock A differentially private text perturbation method using a regularized mahalanobis metric.
\newblock {\em arXiv preprint arXiv:2010.11947}.

\bibitem[Yue et~al., 2021]{yue2021differential}
Yue, X., Du, M., Wang, T., Li, Y., Sun, H., and Chow, S.~S. (2021).
\newblock Differential privacy for text analytics via natural text sanitization.
\newblock {\em arXiv preprint arXiv:2106.01221}.

\end{thebibliography}

\end{document}